\documentclass[conference]{IEEEtran}

\usepackage{amsthm}

\usepackage{subcaption}

\usepackage{cite}
\ifCLASSINFOpdf
\usepackage[pdftex]{graphicx}
\else
\fi

\usepackage{tikz}
\usetikzlibrary{positioning, shapes, arrows.meta}
\usepackage{url}
\usepackage{booktabs}

\begin{document}
\title{Towards a Characterisation of Monte-Carlo Tree Search Performance in Different Games}

% \author{
% \IEEEauthorblockA{anon anon\\
% Anona\\
% Anona\\
% Email: anon@anon}
% \and
% \IEEEauthorblockA{anon anon\\
% Anona\\
% Anona\\
% Email: anon@anon}
% \and
% \IEEEauthorblockA{anon anon\\
% Anona\\
% Anona\\
% Email: anon@anon}}

% \author{Anon}


\author{\IEEEauthorblockN{Dennis J.N.J. Soemers, Guillaume Bams, Max Persoon, Marco Rietjens,\\
Dimitar Sladi{\'c}, Stefan Stefanov, Kurt Driessens, and Mark H.M. Winands}\IEEEauthorblockA{\textit{Department of Advanced Computing Sciences, Maastricht University}\\ Email: \texttt{\{dennis.soemers,kurt.driessens,m.winands\}@maastrichtuniversity.nl}} \texttt{\{g.bams,max.persoon,mjh.rietjens,d.sladic,sp.stefanov\}@student.maastrichtuniversity.nl}
% \IEEEauthorblockA{\IEEEauthorrefmark{2}Twentieth Century Fox, Springfield, USA\\
% Email: homer@thesimpsons.com}
% \IEEEauthorblockA{\IEEEauthorrefmark{3}Starfleet Academy, San Francisco, California 96678-2391\\
% Telephone: (800) 555--1212, Fax: (888) 555--1212}
% \IEEEauthorblockA{\IEEEauthorrefmark{4}Tyrell Inc., 123 Replicant Street, Los Angeles, California 90210--4321}
}


\maketitle

\newcommand{\refalgorithm}[1]{Algorithm~\ref{#1}}
\newcommand{\refassumption}[1]{Assumption~\ref{#1}}
\newcommand{\refdefinition}[1]{Definition~\ref{#1}}
\newcommand{\refequation}[1]{Equation~\eqref{#1}}
\newcommand{\reffigure}[1]{\figurename~\ref{#1}}
\newcommand{\refsection}[1]{Section~\ref{#1}}
\newcommand{\refsubsection}[1]{Subsection~\ref{#1}}
\newcommand{\reftable}[1]{Table~\ref{#1}}
\newcommand{\reftheorem}[1]{Theorem~\ref{#1}}

\begin{abstract}

%Over almost two decades of general game playing and other artificial intelligence research, many variants of and enhancement to Monte-Carlo Tree Search (MCTS) have been proposed. 
Many enhancements to Monte-Carlo Tree Search (MCTS) have been proposed over almost two decades of general game playing and other artificial intelligence research.
However, our ability to characterise and understand which variants work well or poorly in which games is still lacking. This paper describes work on an initial dataset that we have built to make progress towards such an understanding: 268,386 plays among 61 different agents across 1494 distinct games. We describe a preliminary analysis and work on training predictive models on this dataset, as well as lessons learned and future plans for a new and improved version of the dataset.

\end{abstract}


% \fi
\IEEEpeerreviewmaketitle

\section{Introduction}

Monte-Carlo Tree Search (MCTS) \cite{Kocsis_2006_Bandit,Coulom_2007_MCTS} is a commonly used search algorithm for automated game playing. Research from general game playing (GGP) \cite{Pitrat_1968_GGP}, as well as artificial intelligence (AI) for games more broadly, has produced dozens if not hundreds of enhancements to and variants of MCTS \cite{Swiechowski_2022_MCTS}. The levels of playing strength of (modifications to) agents and algorithms are often assessed by averaging or otherwise aggregating over empirical results of experiments in sets of (rarely more than 10-30) different games. Such assessments can provide insights into the expected performance levels of algorithms across sets of games (or other domains), but it is often challenging to gain insights that let us \emph{characterise} the types of games in which specific algorithms or components may be expected to perform particularly well or poorly. 

There have been some analyses of how certain abstract properties of game trees (e.g., the presence or absence of \emph{traps}) \cite{Ramanujan_2010_Adversarial,Ramanujan_2010_Understanding,Ramanujan_2011_Synthetic,Nguyen_2024_Lookahead} affect the performance of MCTS---often using synthetic game trees. However, these insights are not always easy to translate to non-synthetic games, and the analyses tend to be focused on a single instance of MCTS---the canonical UCT \cite{Kocsis_2006_Bandit}---rather than exploring the wide variety of variants that have been proposed over close to two decades of research.

This paper describes our initial attempt at constructing (\refsection{Sec:Dataset}) and analysing (\refsection{Sec:Analysis}) a dataset aimed at making progress towards characterising and understanding which MCTS variants work well or poorly in which (types of) games. The dataset consists of outcomes of 268,386 plays between 61 different agents across 1494 distinct games. Each game is represented by 809 features \cite{Piette_2021_Concepts}, which e.g. machine learning models may use to uncover relationships between properties of games and the relative performance levels of different MCTS variants. Models are trained to predict outcomes between pairs of MCTS agents, and SHAP values \cite{Lundberg_2017_Shap} are used to uncover insights from the features these models use.

%\refsection{Sec:Dataset} provides details on the initial dataset that we have generated. This is followed by a preliminary analysis of the dataset, described in \refsection{Sec:Analysis}. Finally, \refsection{Sec:Conclusion} concludes the paper, with a focus on the lessons that we have learned and will use to construct a new and improved version of the dataset that we plan to publish in the near future.

\section{Dataset} \label{Sec:Dataset}

With a large repository of distinct games, and a consistent set of hundreds of features \cite{Piette_2021_Concepts} to represent these games, Ludii \cite{Piette_2020_Ludii} is an ideal framework to generate a dataset for a large-scale study of the behaviours of different game playing algorithms across different games. We selected and implemented various MCTS agents (\refsubsection{Subsec:Agents}), matched them up against each other in a broad range of different games in Ludii (\refsubsection{Subsec:Games}), and recorded outcomes of all the plays between them (\refsubsection{Subsec:Outcomes}). \reftable{Table:ExampleRowsDataset} illustrates what a small selection of rows from the full dataset looks like.

 \begin{table*}[t]
 \caption{Five example rows from our dataset. Note that there can be multiple rows that are identical except for (or even including) the \textbf{utilities} column, as the same agents playing the same game multiple times may not always lead to the same outcome. The order of agents also matters, as this is the order in which player colours in a game are assigned to the agents.}
 \vspace{-9pt}
 \label{Table:ExampleRowsDataset}
 \begin{center}
 \begin{tabular}{@{}llrrrcr@{}}
 \toprule
 & & \multicolumn{4}{c}{Game Features (809 columns)} & \\
 \cmidrule(lr){3-6}
 \textbf{Game} & \textbf{Agents} & \textbf{Stochastic} & \textbf{AlquerqueBoard} & \textbf{ChessBoard} & $\dots$ & \textbf{Utilities}  \\
 \midrule
 Adugo            & (\textit{UCB1Tuned}-$\sqrt{2}$-\textit{Random0}, \textit{ProgressiveHistory-$0.6$-NST})   & 0    & 1 & 0      & $\dots$   & (1;-1) \\
 Adugo            & (\textit{UCB1Tuned}-$\sqrt{2}$-\textit{Random0}, \textit{ProgressiveHistory-$0.6$-NST})   & 0    & 1 & 0      & $\dots$   & (0;0) \\
 Adugo            & (\textit{ProgressiveHistory-$0.6$-NST}, \textit{UCB1Tuned}-$\sqrt{2}$-\textit{Random0})   & 0    & 1 & 0      & $\dots$   & (-1;1) \\
 Chess            & (\textit{UCB1Tuned}-$0.1$-\textit{MAST}, \textit{UCB1-$0.6$-Random4})   & 0    & 0 & 1      & $\dots$   & (1;-1) \\
 Backgammon            & (\textit{UCB1GRAVE}-$0.1$-\textit{MAST}, \textit{Random})   & 1    & 0 & 0      & $\dots$   & (1;-1) \\
 \bottomrule
 \end{tabular}
 \vspace{-16pt}
 \end{center}
 \end{table*}

\subsection{Agents} \label{Subsec:Agents}

For MCTS-based agents, we selected four different strategies for implementing the algorithm's \textit{selection} phase, three different values for the ``exploration constant'' $C$ used in all of the selection strategies, and five different strategies for the \textit{play-out} phase. These can all be mixed, resulting in a total of $4\times3\times5=60$ different MCTS agents. Finally, we include a Random agent, which plays uniformly at random, for a total of $61$ agents. More concretely, for the MCTS-based agents:
\begin{itemize}
    \item We consider four selection strategies: \textit{UCB1} \cite{Auer_2002_Finite,Kocsis_2006_Bandit}, \textit{UCB1GRAVE} (GRAVE \cite{Cazenave_2015_GRAVE} with $ref=100$ and $bias=10^{-6}$ for its hyperparameters, including a UCB1-style exploration term), \textit{Progressive History} \cite{Nijssen_2011_Enhancements} (with $W = 3$ for its bias influence hyperparameter), and \textit{UCB1Tuned} \cite{Auer_2002_Finite} (using an upper bound on variance of $1$, as Ludii maps losses and wins to the $[-1, 1]$ range).
    \item We consider three values for the exploration constant $C$ (which is a shared hyperparameter among all considered selection strategies): $0.1$, $0.6$, and $\sqrt{2}$.
    \item We consider five different play-out strategies. The first three---\textit{Random0}, \textit{Random4}, and \textit{Random200}---select play-out moves uniformly at random, terminating play-outs early (if they did not yet reach terminal states) after 0, 4, and 200 play-out moves, respectively. The final two play-out strategies are MAST \cite{Finnsson_2008_Simulation} and NST \cite{Tak_2012_NGrams} (both using an $\epsilon$-greedy strategy with $\epsilon = 0.1$, and $n$-grams up to size $n=3$). Both MAST and NST use maximum play-out durations of 200 moves.
\end{itemize}

All MCTS agents re-use relevant parts of search trees built during searches for earlier states when running a new search for a later state. Tree-wide action statistics (for e.g. MAST or NST, but not GRAVE which collects action statistics per node) are decayed by a factor of $0.6$ when re-using data from older searches. Every agent uses $1$ second of processing time per move. Taking into account the setup of our hardware (nodes with 128 cores and 224 GiB of RAM available for jobs) and memory requirements of Ludii (with 7GiB RAM per game, possibly running multiple plays of the same game in parallel, being sufficient for even the largest games), we use 4 cores per search process, using \textit{tree parallelisation} \cite{Chaslot_2008_Parallel} for MCTS.

Many other enhancements and variants of (components of) MCTS \cite{Swiechowski_2022_MCTS} could also have been interesting to include in the study, but including many more was not deemed feasible given the available computation resources. This choice of MCTS variants was made based on their prominence in the GGP literature \cite{Sironi_2019_Thesis}. Any techniques that require offline training time or game-specific heuristics were excluded from consideration, as were techniques that are only applicable to or specifically designed for small subsets of games.

\subsection{Games} \label{Subsec:Games}

Our dataset includes matches played in a total of 1494 different games (i.e., sets of rules by which players may play), all taken from the official Ludii repository. This includes games with one, two, and more than two players. It includes deterministic as well as stochastic games. Games that are classified as ``deduction puzzles'' in Ludii (such as Sudoku) were excluded, as the MCTS-based agents considered in this paper all tend to perform poorly on them. Imperfect-information games were also excluded. Finally, six additional games were specifically excluded because experiments using them in Ludii were often found to take prohibitive amounts of time in prior research, whilst also producing uninteresting results due to consistently poor performance of MCTS in them: \textit{Chinese Checkers}, \textit{Li'b al-'Aqil}, \textit{Li'b al-Gashim}, \textit{Mini Wars}, \textit{Pagade Kayi Ata (Sixteen-handed)}, and \textit{Taikyoku Shogi}.

Every game is represented by a vector of 809 features based on the work of Piette et al. \cite{Piette_2021_Concepts}.\footnote{All features are documented on: \url{https://ludii.games/searchConcepts.php}} Most features are binary, but there are also numeric features, some of which can take on an arbitrary range of values. Features range from highly abstract (e.g., \textit{is the game stochastic?}) to highly specific (e.g., \textit{does this game involve capturing pieces by hopping over them?}). This includes ``play-out concepts,'' which are features of which the values can only be computed (or approximated) from game-play traces. Examples include the frequencies with which certain types of moves (e.g., hopping or stepping) are selected, or statistics relating to the outcomes of plays, such as the percentage of plays that end in a draw. We computed these feature values from play between random agents. 
Similar features computed from play between stronger agents would also add valuable information, but requires significantly more computation time.
%In contrast to computing such features from play between stronger agents (e.g., the MCTS-based agents we wish to study), statistics from random play are cheap to collect, and may be thought of as genuinely being properties of the games alone rather than (also) being properties of the agents themselves.

\subsection{Outcomes} \label{Subsec:Outcomes}

Ideally, we would thoroughly evaluate each of the 60 MCTS-based agents in each of the 1494 different games, against (or with, in cooperative games) any possible combination of all 61 (including \textit{Random}) agents. Every such matchup would also ideally be run many times, because outcomes can vary due to stochasticity in the agents themselves as well as some games' rules. Such an exhaustive evaluation was infeasible given our resources, and as an approximation of this, we scheduled jobs for the following matchups:
\begin{itemize}
    \item For $1$-player games, each of the 60 MCTS-based agents simply plays those games by itself.
    \item For $k$-player games with $k \geq 2$, each of the 60 MCTS-based agents was matched up against $k - 1$ copies of the \textit{Random} agent, and also against a random selection of $k - 1$ out of the 60 MCTS-based agents (including possible mirror matchups in $2$-player games). Note that this is a substantial simplification, as we would ideally evaluate against \textit{all combinations} of $k - 1$ other agents, rather than just a single combination of them.
\end{itemize}
Every such job was scheduled to run $10 \times k$ plays per matchup (with $k$ being the number of players in a game). The order in which the jobs were scheduled was randomised (because we expected to exhaust our computation budget of approximately 500K core-hours before being able to complete them all), and ultimately completed a sample of approximately 10\% of the scheduled jobs. This adds up to 268,386 plays, spread out over many different matchups from a pool of 61 different agents in 1494 distinct games. All agents played some games, matched up against multiple different opponents, and all games were played by at least some different agents, but there is no agent that has played all games against all possible opponents, and there is no game (except for single-player games) that has been played by every possible combination of agents.

For every play, we recorded the outcome in the form of utility scores for each player. Utility scores range from $-1$ to $1$, where $-1$ corresponds to the worst possible outcome in a game (e.g., losing a two-player zero-sum game, or getting the fifth rank in a five-player game), and $1$ corresponds to the best possible outcome (e.g., sole possession of first rank).

\section{Preliminary Analysis} \label{Sec:Analysis}

For a preliminary analysis of the dataset and its potential to help us correlate the playing strength of different variants of MCTS to properties of games, we trained and evaluated a variety of predictive models. 

\subsection{Problem Definition \& Data Preprocessing}

Early on in this research, we settled on the decision to remove any game that is not a zero-sum two-player game from the dataset. In principle, such games would of course also be interesting to retain for analysis. However, they make for a complicating factor in terms of defining consistent input and target output pairings. The vast majority of games in Ludii are in fact two-player zero-sum games, and for these games, a natural way to phrase the prediction problem of interest is to provide a pair of agents (plus game feature vector) as input, and require a model to produce a prediction of relative playing strength (e.g., expected utility for the first of two input agents) as output. This problem definition works for the majority of games in the dataset, but not for games with fewer or more than two players, which would have fewer or more inputs and outputs. Furthermore, the issue that we only have sparse coverage of all the possible data is significantly more severe in games with more players, as the number of combinations of $k$ agents that can be taken, with replacement, from a pool of $61$---the number of matchups of agents we can construct in a $k$-player game---rapidly increases as $k$ increases: $\frac{(61 + k - 1)!}{(k! (61 - 1)!)}$. Hence, getting a sufficient amount of data to make learning feasible for this part of the population is an even greater challenge than it already is for just two-player games.

After settling on the decision to focus on predictions of relative playing strength between pairs of agents in zero-sum games, data on mirror matchups also became of limited use, as an agent playing against itself is always expected to get a utility of $0$. While such data may in principle still provide relevant information on the games themselves (e.g., it might reveal that either player may have a strong advantage in the game), this is not something that we expect the models used in our preliminary analysis to be able to exploit, and therefore we also removed this data. Similarly, plays between an MCTS-based agent and the Random agent may be more likely to reveal valuable information about the game (e.g., is it a highly stochastic or complex game when MCTS fails to outperform random?) than about either one of the agents. Again, we do not expect any of the models considered in our preliminary analysis (which look at one instance at a time) to be able to leverage such information, so we remove any plays involving the Random agent from the dataset. In the end, our dataset has 118,205 plays remaining (down from 268,386), over 1376 unique games (down from 1494). From this dataset, 136 out of the 809 game feature columns were removed because they had exactly the same value for every single row, and 128 columns were removed because they contained missing values. Rows for the same ordered pair of agents in the same game (e.g., the first two rows in the example of \reftable{Table:ExampleRowsDataset}) were merged by averaging over the utilities, as we are interested in predicting expected outcomes. The prediction target is the utility of the first agent, and the utility of the second agent is removed from the data (it is always the complement of the prediction target under the zero-sum assumption). The pair of agents taking part in each row is split up into six columns: three for the first, and three for the second agent, describing the \textit{selection} strategy, \textit{exploration} constant, and \textit{play-out} strategy used by that agent. The columns for \textit{selection} and \textit{play-out} strategies are further split into one-hot encodings.

\subsection{Model Training \& Analysis}

 \begin{table}[t]
 \caption{Mean ($\pm$ standard deviation) of root mean squared errors (RMSE) and mean absolute errors (MAE) from 1376-fold cross validation, for three models.}
 \vspace{-8pt}
 \label{Table:RMSEs}
 \begin{center}
 \begin{tabular}{@{}lrr@{}}
 \toprule
 \textbf{Model} & \textbf{RMSE ($\pm$ std)} & \textbf{MAE  ($\pm$ std)}  \\
 \midrule
 Dummy            & $0.640$ $(\pm 0.267)$ & $0.585$ $(\pm 0.276)$ \\
 Decision Tree            & $0.544$ $(\pm 0.307)$ & $0.465$ $(\pm 0.290)$ \\
 Random Forest            & $0.491$ $(\pm 0.258)$ & $0.434$ $(\pm 0.248)$ \\
 \bottomrule
 \end{tabular}
 \vspace{-16pt}
 \end{center}
 \end{table}

Using default settings from Scikit-learn \cite{Pedregosa_2011_ScikitLearn} version 1.3.2 (unless noted otherwise), we trained dummy regressor models (always predicting the mean of agent 1's utility across all training data), Decision Tree regressors \cite{Breiman_1984_CART} (with maximum tree depths of 10), and Random Forest regressors \cite{Breiman_2001_RandomForests} (with maximum tree depths of 10), for the task of predicting the utility of the first agent for each row of the dataset. 

\reftable{Table:RMSEs} shows the mean and standard deviations in root mean squared errors (RMSEs) and mean absolute errors (MAEs) from a 1376-fold cross validation of the models, where each fold contains all the data of one of the 1376 unique games. We observe some ability to learn meaningful patterns in the data, given the improvements in performance from the Decision Tree and Random Forest over the Dummy regressor, but there also still appears to be room for improvement.

\begin{figure}[t]
    \centerline{\includegraphics[width=\columnwidth]{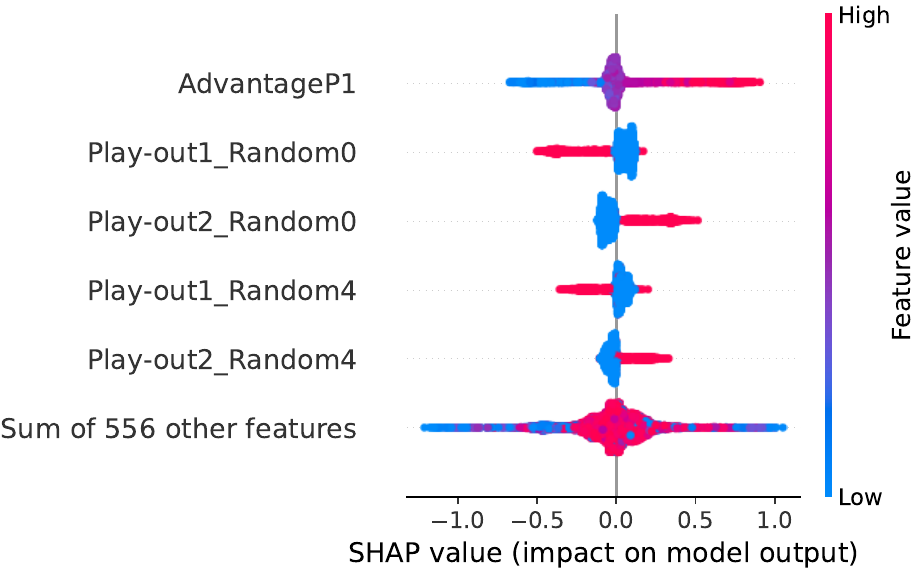}}
    \caption{SHAP \cite{Lundberg_2017_Shap} value estimates of feature importance for the five top individual contributors (plus all other features together), for a Random Forest. Red (blue) means a high (low) feature value. Right (left) means a strong positive (negative) impact on predicted utility for the first agent.}
    \label{Fig:ShapBeeswarm_1}
    \vspace{-12pt}
\end{figure}

\reffigure{Fig:ShapBeeswarm_1} depicts the influence of the topmost five features on model predictions (i.e., on predicted performance of the first agent) as estimated by SHAP \cite{Lundberg_2017_Shap}, for a Random Forest trained on the full dataset. The top row shows that player 1 having a strong advantage (or disadvantage) against player 2---when measured from purely random play-outs---is also used by the model as a predictor of strong (or weak) performance for an MCTS-based player 1---regardless of which variant of MCTS it is. While this does not help us better characterise MCTS performance, it is interesting to see how well estimates of player 1 advantage from random play-outs appear to translate to non-random play. Rows 2--5 show that either player using a random play-out strategy with early termination after $0$ or $4$ moves tends to be a strong predictor for poor performance from that agent, irrespective of the game being played. This was to be expected, as we do not use any handcrafted or trained state evaluation heuristics or functions.

\section{Conclusion and Future Research} \label{Sec:Conclusion}

We have described the construction of our initial dataset of 268,386 plays between 61 different agents across 1494 distinct games, with the aim of making progress towards the characterisation of types of games in which different variants of MCTS work well or poorly. Initial work on training predictive models and analysing them using SHAP values gives some interesting insights. The advantage of one player over another in purely random play appears to still be a surprisingly reliable predictor of outcomes between MCTS agents. We have also learned valuable lessons concerning the construction of the dataset, which we will account for when building a new and improved dataset---planned to be published in the near futurer---before continuing with more elaborate analyses. Play-out strategies that terminate play-outs early perform poorly (in the absence of state evaluation functions), and should be removed so as not to waste further computation time on them. Matching MCTS agents against Random agents may provide useful information about the games themselves, but likely not about the agents involved. Therefore, the proportion of games played against Random agents should be substantially reduced---it is likely only useful to have one such matchup per game, rather than per MCTS agent. Furthermore, it is expected that a more thorough analysis, looking at combinations of features rather than one at a time, will help to make further progress towards our goal.

\section*{Acknowledgment} 
This research was carried out on the Dutch national e-infrastructure with the support of SURF Cooperative (grant no. EINF-4028). This paper is part of the project “Evaluation of Trained AIs for General Game Playing” (with project number EINF-4028) of the research programme Computing Time on National Computer Facilities, which is (partly) financed by the Dutch Research Council (NWO).



\bibliographystyle{IEEEtran}
\bibliography{Dennis-Soemers-BibABBREV}

\begin{thebibliography}{10}
\providecommand{\url}[1]{#1}
\csname url@samestyle\endcsname
\providecommand{\newblock}{\relax}
\providecommand{\bibinfo}[2]{#2}
\providecommand{\BIBentrySTDinterwordspacing}{\spaceskip=0pt\relax}
\providecommand{\BIBentryALTinterwordstretchfactor}{4}
\providecommand{\BIBentryALTinterwordspacing}{\spaceskip=\fontdimen2\font plus
\BIBentryALTinterwordstretchfactor\fontdimen3\font minus \fontdimen4\font\relax}
\providecommand{\BIBforeignlanguage}[2]{{%
\expandafter\ifx\csname l@#1\endcsname\relax
\typeout{** WARNING: IEEEtran.bst: No hyphenation pattern has been}%
\typeout{** loaded for the language `#1'. Using the pattern for}%
\typeout{** the default language instead.}%
\else
\language=\csname l@#1\endcsname
\fi
#2}}
\providecommand{\BIBdecl}{\relax}
\BIBdecl

\bibitem{Kocsis_2006_Bandit}
L.~Kocsis and C.~Szepesv{\'a}ri, ``Bandit based {M}onte-{C}arlo planning,'' in \emph{Mach. Learn.: ECML 2006}, ser. LNCS, J.~F{\"u}rnkranz, T.~Scheffer, and M.~Spiliopoulou, Eds.\hskip 1em plus 0.5em minus 0.4em\relax Springer, Berlin, Heidelberg, 2006, vol. 4212, pp. 282--293.

\bibitem{Coulom_2007_MCTS}
R.~Coulom, ``Efficient selectivity and backup operators in {M}onte-{C}arlo tree search,'' in \emph{Computers and Games}, ser. LNCS, H.~J. van~den Herik, P.~Ciancarini, and H.~H. L.~M. Donkers, Eds., vol. 4630.\hskip 1em plus 0.5em minus 0.4em\relax Springer Berlin Heidelberg, 2007, pp. 72--83.

\bibitem{Pitrat_1968_GGP}
J.~Pitrat, ``Realization of a general game-playing program,'' in \emph{IFIP Congress (2)}, 1968, pp. 1570--1574.

\bibitem{Swiechowski_2022_MCTS}
M.~{\'S}wiechowski, K.~Godlewski, B.~Sawicki, and J.~Ma{\'n}dziuk, ``Monte {C}arlo tree search: A review of recent modifications and applications,'' \emph{Artificial Intell. Review}, vol.~56, pp. 2497--2562, 2022.

\bibitem{Ramanujan_2010_Adversarial}
R.~Ramanujan, A.~Sabharwal, and B.~Selman, ``On adversarial search spaces and sampling-based planning,'' in \emph{Proc. Int. Conf. Automated Planning and Scheduling}, vol.~20, no.~1, 2010, pp. 242--245.

\bibitem{Ramanujan_2010_Understanding}
------, ``Understanding sampling style adversarial search methods,'' in \emph{Proc. 26th Conf. Uncertainty in Artificial Intell.}, 2010, pp. 474--483.

\bibitem{Ramanujan_2011_Synthetic}
------, ``On the behaviour of {UCT} in synthetic search spaces,'' in \emph{ICAPS 2011 Workshop on Monte-Carlo Tree Search: Theory and Appl.}, 2011.

\bibitem{Nguyen_2024_Lookahead}
K.~P.~N. Nguyen and R.~Ramanujan, ``Lookahead pathology in {M}onte-{C}arlo tree search,'' in \emph{Proc. Int. Conf. Automated Planning and Scheduling}, 2024, accepted.

\bibitem{Piette_2021_Concepts}
{\'E}.~Piette, M.~Stephenson, D.~J. N.~J. Soemers, and C.~Browne, ``General board game concepts,'' in \emph{Proc. 2021 IEEE Conf. on Games}.\hskip 1em plus 0.5em minus 0.4em\relax IEEE, 2021, pp. 932--939.

\bibitem{Lundberg_2017_Shap}
S.~M. Lundberg and S.-I. Lee, ``A unified approach to interpreting model predictions,'' in \emph{Adv. Neural Inf. Proc. Syst. 30}, I.~Guyon, U.~V. Luxburg, S.~Bengio, H.~Wallach, R.~Fergus, S.~Vishwanathan, and R.~Garnett, Eds.\hskip 1em plus 0.5em minus 0.4em\relax Curran Associates, Inc., 2017, pp. 4765--4774.

\bibitem{Piette_2020_Ludii}
{\'E}.~Piette, D.~J. N.~J. Soemers, M.~Stephenson, C.~F. Sironi, M.~H.~M. Winands, and C.~Browne, ``Ludii -- the ludemic general game system,'' in \emph{Proc. 24th Eur. Conf. Artif. Intell.}, ser. Frontiers in Artificial Intell. and Appl., G.~D. Giacomo, A.~Catala, B.~Dilkina, M.~Milano, S.~Barro, A.~Bugarín, and J.~Lang, Eds., vol. 325.\hskip 1em plus 0.5em minus 0.4em\relax IOS Press, 2020, pp. 411--418.

\bibitem{Auer_2002_Finite}
P.~Auer, N.~Cesa-Bianchi, and P.~Fischer, ``Finite-time analysis of the multiarmed bandit problem,'' \emph{Mach. Learn.}, vol.~47, no. 2--3, pp. 235--256, 2002.

\bibitem{Cazenave_2015_GRAVE}
T.~Cazenave, ``{Generalized Rapid Action Value Estimation},'' in \emph{{Proc. 24th Int. Joint Conf. Artificial Intell.}}, Q.~Yang and M.~Woolridge, Eds.\hskip 1em plus 0.5em minus 0.4em\relax AAAI Press, 2015, pp. 754--760.

\bibitem{Nijssen_2011_Enhancements}
J.~A.~M. Nijssen and M.~H.~M. Winands, ``Enhancements for multi-player {M}onte-{C}arlo tree search,'' in \emph{Computers and Games (CG 2010)}, ser. LNCS, H.~J. van~den Herik, H.~Iida, and A.~Plaat, Eds.\hskip 1em plus 0.5em minus 0.4em\relax Springer Berlin Heidelberg, 2011, vol. 6515, pp. 238--249.

\bibitem{Finnsson_2008_Simulation}
H.~Finnsson and Y.~Bj{\"o}rnsson, ``Simulation-based approach to general game playing,'' in \emph{Proc. 23rd AAAI Conf. Artif. Intell.}\hskip 1em plus 0.5em minus 0.4em\relax {AAAI Press}, 2008, pp. 259--264.

\bibitem{Tak_2012_NGrams}
M.~J.~W. Tak, M.~H.~M. Winands, and Y.~Bj{\"o}rnsson, ``N-grams and the last-good-reply policy applied in general game playing,'' \emph{IEEE Trans. Comput. Intell. AI Games}, vol.~4, no.~2, pp. 73--83, 2012.

\bibitem{Chaslot_2008_Parallel}
G.~M. J.-B. Chaslot, M.~H.~M. Winands, and H.~J. van~den Herik, ``Parallel {M}onte-{C}arlo tree search,'' in \emph{Int. Conf. Computers and Games}, ser. LNCS, H.~J. van~den Herik, X.~Xu, Z.~Ma, and M.~H.~M. Winands, Eds., vol. 5131.\hskip 1em plus 0.5em minus 0.4em\relax Springer, Berlin, Heidelberg, 2008, pp. 60--71.

\bibitem{Sironi_2019_Thesis}
C.~F. Sironi, ``{M}onte-{C}arlo tree search for artificial general intelligence in games,'' PhD thesis, Department of Data Science and Knowledge Engineering, Maastricht University, Maastricht, the Netherlands, 2019.

\bibitem{Pedregosa_2011_ScikitLearn}
F.~Pedregosa, G.~Varoquaux, A.~Gramfort, V.~Michel, B.~Thirion, O.~Grisel, M.~Blondel, P.~Prettenhofer, R.~Weiss, V.~Dubourg, J.~Vanderplas, A.~Passos, D.~Cournapeau, M.~Brucher, M.~Perrot, and E.~Duchesnay, ``Scikit-learn: Machine learning in {P}ython,'' \emph{Journal of Machine Learning Research}, vol.~12, pp. 2825--2830, 2011.

\bibitem{Breiman_1984_CART}
L.~Breiman, J.~Friedman, R.~A. Olshen, and C.~J. Stone, \emph{Classification and Regression Trees}.\hskip 1em plus 0.5em minus 0.4em\relax New York, USA: Chapman and Hall/CRC, 1984.

\bibitem{Breiman_2001_RandomForests}
L.~Breiman, ``Random forests,'' \emph{Mach. Learn.}, vol.~45, no.~1, pp. 5--32, 2001.

\end{thebibliography}

\end{document}